\documentclass[journal]{IEEEtran}
\pdfoutput=1
\usepackage[english]{babel}
\usepackage{algorithm,algorithmic}
\usepackage{color}
\usepackage{lineno}
\usepackage{graphicx}
\usepackage{amsfonts}
\usepackage{amssymb}
\usepackage{amsmath}
\usepackage{theorem}

\newcommand{\R}{\mathbb{R}}

\newcommand{\change}{}
\newcommand{\argmin}{\mathop{\mathrm{arg\,min}}}
 
\title{A Plug\&Play P300 BCI Using Information Geometry}
\author{\IEEEauthorblockA{Alexandre Barachant, Marco Congedo}\\
\IEEEauthorblockA{Team ViBS (Vision and Brain Signal Processing), GIPSA-lab, CNRS, Grenoble Universities. France.\\
Email: alexandre.barachant@gmail.com}
}

\renewcommand{\rightmark}{Barachant \MakeLowercase{\textit{et al.}}: 
A Plug\&Play P300 BCI Using Information Geometry}

\begin{document}
\maketitle
\begin{abstract}
This paper presents a new classification methods for Event Related Potentials (ERP) based on an Information geometry framework. Through a new estimation of covariance matrices, this work extend the use of Riemannian geometry, which was previously limited to SMR-based BCI, to the problem of classification of ERPs. As compared to the state-of-the-art, this new method increases performance, reduces the number of data needed for the calibration and features good generalisation across sessions and subjects. This method is illustrated on data recorded with the P300-based game \emph{brain invaders}. Finally, an online and adaptive implementation is described, where the BCI is initialized with generic parameters derived from a database and continiously adapt to the individual, allowing the user to play the game without any calibration while keeping a high accuracy.
\end{abstract}


\section{Introduction}

So far we have conceived a Brain-Computer Interface (BCI) as a learning machine where the classifier is trained in a calibration phase preceding immediately the actual BCI use~\cite{lotte2007review}. Depending on the BCI paradigm and on the efficiency of the classifier, the calibration phase may last from a few to several minutes. Regardless the duration, the very necessity of a calibration session reduces drastically the usability and appealing of a BCI. This is true both for clinically-oriented BCI, where the cognitive skills of patients are often limited and are wasted in the calibration phase, and for healthy users where the “plug\&play” operation is nowadays considered as a minimum requirement for any consumer interfaces and devices. Besides the essential considerations from the user perspective, it appears evident that training the BCI at the beginning of each session and discarding the calibration data at the end is a very inefficient way to proceed. 
The problem we pose here is: can we design a "plug\&play" BCI? Of course, such a goal does not imply that the BCI is not calibrated. What we want to achieve is that the calibration is completely hidden to the user. One possible solution is to initialize the classification algorithm with generic parameters derived from previous sessions and/or previous users and then to adapt continuously these parameters during the experiment in order to reach optimality for the current user in the current session. In order to do so our BCI must possess two key properties, namely:
\begin{enumerate}
\item good \emph{across-subject generalization} and \emph{across-session generalization}, so as to yield appropriate accuracy (albeit sub-optimal) at the beginning of each session and even for the very first session of a BCI naive user.
\item \emph{fast convergence} toward the optimal parameters, that is, the adaptation should be efficient and robust (the adaptation is useless if it is effective only at the end of the session).
\end{enumerate}
The two properties above are not trivial. On one hand, the methods involving spatial filtering enjoy fast convergence, but show a bad generalisation since the subspace projection is very specific to each subject and session. On the other hand, the methods working directly with low level, high dimensional space or features selected space show good generalization but require a lot of data to avoid over-fitting and therefore do not fulfil the property of fast convergence.

We have solved this problem before by working in the native space of covariance matrices and using Riemannian geometry~\cite{barachant_multiclass_2012}. So far this framework was limited to BCIs based on motor imagery and SSVEP since the regular covariance matrix estimation does not take into account the temporal structure of the signals, which is critical for ERP detection. Here we allow the use of our Riemannian geometry framework for ERP-based BCI by introducing a new smart way for building covariance matrices. We show experimentally that our new initialization/adaption BCI framework (based on Riemannian geometry) enjoys both properties 1) and 2). In addition, our method is rigorous, elegant, conceptually simple and computationally efficient. As a by-product the implementation is easier, more stable and compatible with the time constraints of a system operating on-line. Finally, thanks to the advance here presented we are able to propose a universal framework for calibration-less BCIs, that is, the very same signal processing chain can be now applied with minor modifications to any currently existing types of BCI.

This paper is organized as follows. 
Section~\ref{sec:infogeom} introduce the metric, based on information geometry, that we are using in this work.
Section~\ref{sec:bigmethod} presents the classification algorithm and the new covariance matrix estimation method dedicated to the classification of ERPs.
The method is illustrated on the P300 ERP. Results on three datasets are shown in section~\ref{sec:bigresults} and compared to the state-of-the-art methods. Finally, we will conclude and discuss in section~\ref{sec:conclusion}.

\change{
\subsubsection*{Related work}
Over the past years, several attempt to design "plug\&play" BCI have been made. In~\cite{fazli2009subject}, a pure machine-learning solution has been used for this purpose. For each user in a database, a complete classification chain (band-pass filtering, CSP spatial filtering, LDA classifier) is trained and optimized. Then, for a new user, all these classifiers are applied and the results are combined (different solutions are proposed) in order to build a single classification result.
In \cite{kindermans2012bayesian,kindermans2012p300}, an unsupervised adaptation of a generic classifier is proposed to deal with inter-session and inter-subject variability in the context of the P300 speller. The proposed solution makes the use of the P300 constraints and language model to adapt a bayesian classifier and unsure the convergence toward the optimal classifier. This solution is for now the most convincing approach for the P300 speller.
}

\section{A distance for the multivariate normal model}
\label{sec:infogeom}
\subsection{Motivation}
Our goal is to use a method enjoying both properties of fast adaptation and generalisation across subjects and sessions in order to build an efficient adaptive classification chain. As we will see, none of the state-of-the-art methods fulfils these requirement, therefore we have to design a new classification algorithm.

In essence, a classification algorithm works by comparing data to each other or to a prototype of the different classes. The notion of comparison implies the ability to measure dissimilarity between the data, i.e., the definition of a metric. Obviously, the accuracy of the classification is related to the ability of the metric to measure difference between classes for the problem at hand. The definition of the appropriate metric is not trivial and should be accomplished according to the intrinsic nature of the data. This problem is usually bypassed by the feature extraction process, which projects the raw data in a feature space where a well known metric (Euclidean distance, Mahalanobis distance, etc) is effective. When the feature extraction is not able to complete the task, the metric could be embedded within the classification function by using a kernel trick, as in the Support Vector Machine (SVM) classifier. Nevertheless, the problem of choosing the kernel is not obvious and is generally solved by trials and errors, that is, by benchmarking several predefined kernel through a cross-validation procedure. However proceeding this way one loose generalization~\cite{lotte2007review}.

The recent BCI systems follow these approaches, by using a \change{subspace} separation~\cite{gouy2010nonstationary} procedure and a selection of relevant \change{dimensions}~\cite{grosse2008multiclass} before applying the classifier on the extracted features. In these approaches, the weakness of the chain is the subspace separation process, which rely on several assumptions that are more or less fulfilled, depending on the number of classes and the quality of the data. In addition, the \change{subspace} separation gives solutions that are not generalisable between session and subjects.
To overcome these difficulties, the linear \change{subspace} separation could be embedded in the classifier by working directly on the covariance matrices~\cite{farquhar_linear_2009} and using a SVM kernel to classify them~\cite{reuderink_subject-independent_2011}. Again, the choice of the kernel and its parameters is done heuristically which is incompatible with adaptive implementations.

In this context, the definition of a metric adapted to the nature of EEG signals seems to be particularly useful. With such a metric, a very simple distance-based classifier can be used, allowing efficient adaptive implementations, reducing the amount of data needed for calibration (no cross-validation or feature selection) and offering better generalization between subjects and sessions (no source separation or any other spatial filtering).

\subsection{An Information geometry metric}
Thanks to information geometry we are able to define a metric enjoying the sought properties. The information geometry is a field of information theory where the probability distributions are taken as point of a Riemannian manifold. This field has been popularised by S. Amari who has widely contributed to establish the theoretical background~\cite{amary}. Information geometry has today reached its maturity and find an increasing number of practical applications, such as radar image processing~\cite{barbaresco_innovative_2008}, diffusion tensor imaging~\cite{fletcher_principal_2004} and image and video processing~\cite{tuzel_pedestrian_2008}. The strength of the information geometry is to give a natural choice for an information-based metric according to the probability distribution family: the \emph{Fisher Information}~\cite{amary}. Therefore, each probability distribution belongs to a Riemannian manifold with a well established metric leading to the definition of a distance between probability distributions. 

In this work, EEG signals are assumed to follow a multivariate normal distribution  with a zero mean (this is the case for EEG signals that are  high-pass filtered), which is a common assumption~\cite{congedo2008blind}. More specifically, it has been shown that EEG signal are block stationary and can be purposefully modelled by second order statistics~\cite{comon2010handbook}.
\change{
Let us denote by $\mathbf{X} \in \R^{C \times N}$ an EEG trial\footnote{An epoch of EEG signal we want to classify.}  , recorded over $C$ electrodes and $N$ time samples. We assume that this trial follow a multivariate normal model with zero mean and with a covariance matrix $\mathbf{\Sigma} \in \R^{C \times C}$, i.e. 
\begin{equation}
\mathbf{X} \sim \mathcal{N}(\mathbf{0},\mathbf{\Sigma})
\end{equation}
}
The application of information geometry to the multivariate model with zero mean provides us with a metric that can be used to compare directly covariance matrices of EEG segments.

\subsection{Distance}
Let us denote two EEG trials $\mathbf{X}_1\sim \mathcal{N}(\mathbf{0},\mathbf{\Sigma}_1)$ and $\mathbf{X}_2\sim \mathcal{N}(\mathbf{0},\mathbf{\Sigma}_2)$. Their Riemannian distance according to information geometry is given by~\cite{barbaresco_innovative_2008}
\begin{equation}
\label{eq:Rgeodistance}
\delta_R(\mathbf{\Sigma}_1,\mathbf{\Sigma}_2) 
= 
\Vert \mathrm{log} \left( \mathbf{\Sigma}_1^{-1/2} \mathbf{\Sigma}_2 \mathbf{\Sigma}_1^{-1/2} \right) \Vert_F
=
\left[ \sum_{c=1}^{C} \log^2 \lambda_c \right]^{1/2},
\end{equation}
where $\lambda_c, c=1\ldots C$ are the real eigenvalues 
of $\mathbf{\Sigma}_1^{-1/2} \mathbf{\Sigma}_2 \mathbf{\Sigma}_1^{-1/2}$ and $C$ the number of electrodes.
 Note that this distance is also called \emph{Affine-invariant}~\cite{arsigny2007geometric} when it is obtained through the algebraic properties of the spaces of symmetric positive definite matrices rather than through the information geometry work-flow.
This distance has two important properties of invariance. First, the distance is invariant by inversion, i.e., 
\begin{equation}
\label{eq:invariance1}
\delta_R(\mathbf{\Sigma}_1,\mathbf{\Sigma}_2) = \delta_R(\mathbf{\Sigma}^{-1}_1,\mathbf{\Sigma}^{-1}_2).
\end{equation}

Second, the distance is invariant by congruent transformation, meaning that for any invertible matrix $\mathbf{V} \in \mathcal{G}l(C)$ we have
\begin{equation}
\label{eq:invariance2}
\delta_R(\mathbf{\Sigma}_1,\mathbf{\Sigma}_2) = \delta_R(\mathbf{V}^T\mathbf{\Sigma}_1\mathbf{V},\mathbf{V}^T\mathbf{\Sigma}_2\mathbf{V}).
\end{equation}
This latter property has numerous consequences on EEG signal. The most remarkable can be formulated as it follows:

\emph{
Any linear operation on the EEG signals that can be modelled by a $C \times C$ invertible matrix $\mathbf{V}$ has no effect on the Riemannian distance.}

Such a class of transformations include rescaling and normalization of the signals, electrodes permutations, whitening, spatial filtering or source separation, etc. In essence, manipulating covariance matrices with Eq.(\ref{eq:Rgeodistance}) is equivalent to working in the optimal source space of dimension $C$.

\subsection{Geometric mean}
\label{sec:geomean}
The Riemannian geometric mean of $I$ covariance matrices (denoted by $\mathfrak{G}(.)$), also called Fr\'echet mean, is defined as the matrix minimizing the sum of the squared Riemannian distances~\cite{pennec2006riemannian}, i.e.,

\begin{equation}
\mathfrak{G} \left( \mathbf{\Sigma}_1,\ldots,\mathbf{\Sigma}_I \right) = \argmin_{\mathbf{\Sigma} \in P(n)} 
\sum_{i=1}^{I} 
\delta_R^2 \left( \mathbf{\Sigma},\mathbf{\Sigma}_i \right).
\label{eq:geo_mean}
\end{equation}
There is no closed form expression for this mean, however a gradient descent in the manifold can be used in order to find the solution~\cite{pennec2006riemannian}. A matlab implementation is provided in a Matlab toolbox\footnote{http://github.com/alexandrebarachant/covariancetoolbox}.
Note that the geometric mean inherits the properties of invariance from the distance.

\subsection{Geodesic}
The geodesic is defined as the shortest curve between two points \change{(in this work, covariance matrices)} of the manifold~\cite{pennec2006riemannian}. The geodesic according to the Riemannian metric is given by
\begin{equation}
\label{eq:geodesicmvn}
\mathbf{\Gamma}(\mathbf{\Sigma}_1,\mathbf{\Sigma}_2,t)
= 
\mathbf{\Sigma}_1^{\frac{1}{2}}
\left( 
\mathbf{\Sigma}_1^{-\frac{1}{2}}\mathbf{\Sigma}_2\mathbf{\Sigma}_1^{-\frac{1}{2}}
\right)^t 
\mathbf{\Sigma}_1^{\frac{1}{2}},
\end{equation}
with $t \in [0:1]$ a scalar. Note that the geometric mean of two points is equal to the geodesic at $t=0.5$~\cite{moakher_differential_2005}, i.e.,  
\begin{equation}
\mathfrak{G} \left( \mathbf{\Sigma}_1,\mathbf{\Sigma}_2 \right)
=
\mathbf{\Gamma}(\mathbf{\Sigma}_1,\mathbf{\Sigma}_2,0.5)
\end{equation}

\change{
Moreover, the geodesic could be seen as a geometric interpolation between the two covariance matrices $\mathbf{\Sigma}_1$ and $\mathbf{\Sigma}_2$. Its Euclidean equivalent is the well known linear interpolation $  (1 - t ) \mathbf{\Sigma}_1 + t\mathbf{\Sigma}_2$.
}

\section{Method}
\label{sec:bigmethod}

\subsection{Classification of covariance matrices}
Using the metric defined by the information geometry we are now able to efficiently compare EEG trials using only their covariance matrices. 

The direct classification of covariance matrices has proved very useful for single trial classification in the context of SMR-based BCI~\cite{barachant_multiclass_2012,barachant_neurocomp_2013,reuderink_subject-independent_2011} and more generally in every EEG application where the relevant features are based on the spatial patterns of the signal. A similar approach, working on the power spectral density rather than the covariance matrix, has been developed for sleep state detection~\cite{li2009eeg,li2012electroencephalogram}. 

According to the property of invariance of the distance, the Riemannian metric allows to bypass the source separation step by exploiting directly the spatial information contained in the covariance matrix. While the state-of-the-arts methods rely on the classification of the log-variances in the source space, the Riemannian metric gives access to the same information in the sensor space, leading to a better generalization of the trained classifier.

The Minimum Distance to Mean (MDM) algorithm is one of the simplest classification algorithm based on distance comparison. Used with the Riemannian distance, it is a powerful and efficient way to build a an online BCI system~\cite{barachant2011brain}. Given a training set of EEG trials $\mathbf{X}_i\sim \mathcal{N}(\mathbf{0},\mathbf{\Sigma}_i)$ and the corresponding labels $y_i \in \lbrace 1 :  N_c \rbrace$, the classifier training consist in the estimation of a mean covariance matrix for each of the $N_c$ classes, 
\begin{equation}
\mathbf{\bar{\Sigma}}_{K} = \mathfrak{G} \left( {\mathbf{\Sigma}}_i \vert y_i = K \right),
\label{eq:geomean}
\end{equation}
where $K \in [1:N_c]$ is the class label and $\mathfrak{G}(.)$ is the Riemannian geometric mean defined in section~\ref{sec:geomean}. In essence, the geometric mean represents the expected distribution that will follows a trial belonging to class $K$. 

Then, the classification of a new trial $\mathbf{X} \sim \mathcal{N}(\mathbf{0},\mathbf{\Sigma})$ of an unknown class $y$ is simply achieved by looking at the minimun distance to each mean, 
\begin{equation}
\hat{y} = \argmin_K \delta_R(\mathbf{\bar{\Sigma}}_{K},\mathbf{\Sigma}).
\label{eq:classif}
\end{equation}
Such a classifier defines a set of vorono\"i cells in the manifold.
Alternatively, for binary classification problems, the difference of the two distances could be used as a linear classification score.
The algorithm is extremely simple since it involve only a mean estimation and a distance computation for each class. The full algorithm is given in Algorithm~\ref{algo:distance_classif}.
\begin{algorithm}
\begin{footnotesize}\caption{Minimum Distance to Riemannian Mean}
Input: a set of trials $\mathbf{X}_i$ of $N_c$ different known classes and the corresponding labels $y_i \in \lbrace 1:K \rbrace$.\\
Input: $\mathbf{X}$ an EEG trial of unknown class.\\
Output: $ \hat{y}$ the estimated class of test trial $\mathbf{X}$.
\begin{algorithmic}[1]
\label{algo:distance_classif}
\STATE Compute SCMs of $\mathbf{X}_i$ to obtain $\mathbf{\Sigma}_i$.
\STATE Compute SCM of $\mathbf{X}$ to obtain $\mathbf{\Sigma}$.
\FOR {$K=1$ to $N_c$}
	\STATE $\mathbf{\bar{\Sigma}}_{K} = \mathfrak{G} \left( {\mathbf{\Sigma}}_i \vert y_i = K \right)$, Riemannian geometric mean (\ref{eq:geo_mean}).
\ENDFOR
\STATE $\hat{y} = \argmin_K  \delta_R(\mathbf{\bar{\Sigma}}_{K},\mathbf{\Sigma}) $,  Riemannian distance (\ref{eq:Rgeodistance}).
\RETURN $ \hat{y} $
\end{algorithmic}\end{footnotesize}
\end{algorithm}

\subsection{Classification of covariance matrices in P300-based BCI}
\label{sec:method}
The above procedure is not suitable for classification of ERPs, where the discriminant information is given by the waveform of the potential rather than the spatial distribution of its variance. Indeed, the covariance matrix estimation does not take into account the time structure of the signal, and therefore using only the covariance matrix for the classification leads to the loss of the information carried by the temporal structure of the ERP.

More generally, the prior knowledge of a reference signal, i.e., the expected waveform of the signal due to a physiological process (P300, ErrP, etc.) or induced by a stimulation pattern (SSEP), are rarely taken into account in the signal processing chain.

Our contribution stems from the definition of a specific way of building covariance matrices for ERP data embedding both the spatial and temporal structural information. This new definition, making use of a reference signal, allows the purposeful application of Riemannian geometry for ERP data. Let us define $\mathbf{P1} \in \R^{C \times N}$ the prototyped ERP response, obtained, for example, by averaging several single trial responses \change{of one class, usually the target class,} such as
\begin{equation}
\mathbf{P1} = \frac{1}{\vert \mathcal{I} \vert}\sum_{i \in \mathcal{I}} \mathbf{X}_i.
\label{eq:P1}
\end{equation}
\change{where $\mathcal{I}$ is the ensemble of indexes of the trial belonging to the target class.}
For each trial $\mathbf{X}_i \in \R^{C\times N}$, we build a super trial $\tilde{\mathbf{X}}_i \in \R^{2C\times N}$ by concatenating  $\mathbf{P1}$ and $\mathbf{X}_i$~:
\begin{equation}
\tilde{\mathbf{X}}_i =  \left[ 
\begin{array}{c}
\mathbf{P1} \\ 
\mathbf{X}_i
\end{array} \right].
\label{eq:concatenation}
\end{equation}
These super trials are used to build the covariance matrices on which the classification will be based. The covariance matrices are estimated using the Sample Covariance Matrix (SCM) estimator:
\begin{equation}
\tilde{\mathbf{\Sigma}}_i = \frac{1}{N-1} \tilde{\mathbf{X}}_i \tilde{\mathbf{X}}_i^T.
\label{eq:scm}
\end{equation}

These covariances matrices can be decomposed in three different blocks~: $\mathbf{\Sigma}_{P1}$ the covariance matrix of $\mathbf{P1}$,  $\mathbf{\Sigma}_i$ the covariance matrix of  $\mathbf{X}_i$ and $\mathbf{C}_{P1,X_i}$ the cross-covariance matrix (non-symmetric) between  $\mathbf{P1}$ and  $\mathbf{X}_i$, that is,
\begin{equation}
\tilde{\mathbf{\Sigma}}_i =  \left[ 
\begin{array}{cc}
\mathbf{\Sigma}_{P1} &  \mathbf{C}_{P1,X_i}^T\\ 
\mathbf{C}_{P1,X_i} & \mathbf{\Sigma}_i
\end{array} \right].
\label{eq:covblock}
\end{equation}
$\mathbf{\Sigma}_{P1}$ is the same for all the trials. This block is not useful for the discrimination between the different classes. $\mathbf{\Sigma}_i$ is the covariance matrix of the trial $i$. As explained previously, this block is not sufficient to achieve a high classification accuracy. However, it contains information that can be used in the classification process.
The cross-covariance matrix $\mathbf{C}_{P1,X_i}$ contains the major part of the useful information. If the trial contains an ERP in phase with $\mathbf{\Sigma}_{P1}$, the cross-covariance will be high for the corresponding channels. In other cases (absence of ERP or non phase locked ERP), the cross-covariance will be close to zero, leading to a particular covariance structure that can be exploited by the classification algorithm.

Note that the property of invariance of the metric allows to use a reduced version of $\mathbf{P1}$ after a source separation and a selection of relevant sources if desired.

Using this new estimation, the MDM algorithm can be applied in the exact same manner, using $\tilde{\mathbf{X}}_i$ Eq.(\ref{eq:concatenation}) instead of $\mathbf{X}_i$ in Algorithm~\ref{algo:distance_classif}. In the case of ERP, only two classes are used, target for trials containing an ERP and non-target for the others, leading to the estimation of two mean covariance matrices, denoted $\mathbf{\bar{\Sigma}}_T$ and $\mathbf{\bar{\Sigma}}_{\bar{T}}$,  respectively. For this binary classification task, the classification score for each trial is given by the difference of the Riemannian distance to each class, i.e.,
\begin{equation}
s = \delta_R(\mathbf{\bar{\Sigma}}_{\bar{T}},\mathbf{\Sigma})-\delta_R(\mathbf{\bar{\Sigma}}_{{T}},\mathbf{\Sigma}).
\label{eq:score}
\end{equation}
\subsection{An adaptive implementation}
\label{sec:adaptive}

We employ here a simple adaptation scheme. When new data from the current subject is available, the two subject-specific intra-class mean covariance matrices  $\mathbf{\bar{\Sigma}}^s_T$ and $\mathbf{\bar{\Sigma}}^s_{\bar{T}}$ are estimated and combined with the two generic ones  (computed on a database) $\mathbf{\bar{\Sigma}}^g_T$ and $\mathbf{\bar{\Sigma}}^g_{\bar{T}}$ using a geodesic interpolation:
\begin{equation}
\mathbf{\bar{\Sigma}}_K= \mathbf{\Gamma}(\mathbf{\bar{\Sigma}}^g_K,\mathbf{\bar{\Sigma}}^s_K,\alpha),
\end{equation}
where $\mathbf{\Gamma}$ is the geodesic from $\mathbf{\bar{\Sigma}}^g_K$ to $\mathbf{\bar{\Sigma}}^s_K$  given in Eq.(\ref{eq:geodesicmvn}), with $K$ denoting $T$ ($\bar{T}$) for the target (non-target) class and with $\alpha \in [0:1]$ a parameter settings the strength of the adaptation evolving from 0 at the beginning of the session to 1 at the end of the session. 
This combination is the Riemannian equivalent to $(1-\alpha)\mathbf{\bar{\Sigma}}^g_K + \alpha\mathbf{\bar{\Sigma}}^s_K$ in the euclidean space.
The combined matrices are then used for the classification in replacement of the standard mean covariances matrices in Algorithm~\ref{algo:distance_classif}.
A similar approach has been presented for the adaptation of spatial filters in~\cite{lotte2011regularizing}. 

The adaptation is supervised, i.e., the classes of the trials have to be known. This not a limitation in applications where the target is chosen by the interface itself and not by the user. For other ERP-based BCI such as a P300 speller, adaptation will be somehow slower since only past symbols (that the user has confirmed) can be used. However, an unsupervised adaptation is also possible using clustering of covariance matrices in Riemannian space. Preliminary results (not shown here) show that this approach is equivalent to the supervised one.

\section{Results}
\label{sec:bigresults}
\subsection{Material}
The presented method is illustrated through the detection of P300 evoked potentials on three datasets. These datasets are recorded using the P300-based game \emph{Brain Invaders}~\cite{congedo_brain_invaders}. Brain invaders is inspired from the famous vintage game Space invaders. The goal is to destroy a target alien in a grid of non-target alien using the classical P300 paradigm. After each repetition, i.e., flash of the 6 rows and 6 columns, the most probable alien is destroyed according to the single trial classification scores of the 12 flashes. If the target alien is destroyed, the game switch to the next level. If not, a new repetition begins and its classification scores will be averaged with the scores of the previous repetitions. Therefore, the goal of the game is to complete all levels in a minimum number of repetitions.

\subsection{Preprocessing and State-of-the-art methods}
\change{Motivate the Choice Reference methods}

For each method and dataset, signals are bandpass filtered by a 5th order butterworth filter between 1 and 20 Hz. Filtered signal are then downsampled to 128 Hz and segmented in epochs of 1s starting at the time of the flash.
Two state of the art methods have been implemented for comparison with the proposed method. The first is named XDAWN and is composed by a spatial filtering using the xDAWN method~\cite{rivet2009xdawn} and a regularized LDA~\cite{swlda} for classification. After the estimation of spatial filters, the two best spatial filters are applied on the epochs. The spatially filtered epoch are then downsampled by a factor four and the time sample are aggregated to build a feature vector of size $32\times 2 = 64$. These feature vectors are then classified using a regularized LDA.

The second method is name SWLDA and is composed by a single classification stage using a stepwise LDA~\cite{swlda}. First the signals were downsampled by a factor four and aggregated to build a feature vector of size $32\times C$. These feature vectors are classified using a stepwise LDA~\cite{swlda}. The stepwise LDA performs a stepwise model selection before applying a regular linear discriminant analysis, reducing the number of features used for classification.

The MDM method is directly applied on the epochs of signal, with no other preprocessing besides the frequency band-pass filtering and downsampling.

\subsection{Dataset I}

Dataset I was recorded in laboratory conditions using 10 silver/silver-chloride electrodes (T7-Cz-T8-P7-P3-PZ-P4-P8-O1-O2) and a Nexus amplifier (TMSi, The Netherlands). The signal was sampled at 512 Hz. 23 subjects participated to this experiment for one session composed of a calibration phase of 10 minutes and a test phase of 10 minutes. This dataset is used to evaluate offline performances of the methods in the canonical training-test paradigm.
\newline
\subsubsection{Canonical performances}
\label{sec:canon}
In this section, performances will be evaluated using the classical training-test paradigm. Algorithms are calibrated on the data recorded during the training phase, and applied on the data recorded during the test phase (online). The Area Under ROC Curve (AUC), estimated on the classification scores Eq.(\ref{eq:score}) will be used as criterion to evaluate the effectiveness of the trained classifier. 
Table~\ref{tab:result_training_test} reports these AUC for the 23 subjects and the three evaluated methods.

\begin{table*}[ht]
\centering
\caption{AUC on test data for the 23 subjects of the dataset I for the 3 evaluated methods.}
\label{tab:result_training_test}
\begin{tabular}{|l|cccccccccccc|}
	\hline
	\textbf{Methdod} & \textbf{1} & \textbf{2} & \textbf{3} & \textbf{4} & \textbf{5} & \textbf{6} & \textbf{7} & \textbf{8} & \textbf{9} & \textbf{10} & \textbf{11} & \textbf{12}\\
	\hline
	MDM & 0.89 & 0.93 & 0.97 & 0.88 & 0.95 & 0.90 & 0.65 & 0.65 & 0.87 & 0.97 & 0.94 & 0.86\\
	SWLDA & 0.84 & 0.90 & 0.97 & 0.81 & 0.90 & 0.91 & 0.55 & 0.69 & 0.79 & 0.96 & 0.89 & 0.84\\
	XDAWN & 0.89 & 0.90 & 0.97 & 0.85 & 0.91 & 0.90 & 0.56 & 0.66 & 0.79 & 0.95 & 0.83 & 0.86\\
	\hline
	 & \textbf{13} & \textbf{14} & \textbf{15} & \textbf{16} & \textbf{17} & \textbf{18} & \textbf{19} & \textbf{20} & \textbf{21} & \textbf{22} & \textbf{23} & \textbf{mean (std)}\\
	\hline
	MDM & 0.88 & 0.92 & 0.93 & 0.91 & 0.98 & 0.84 & 0.78 & 0.97 & 0.90 & 0.95 & 0.96 & \textbf{0.89 (0.09)}\\
	SWLDA & 0.87 & 0.91 & 0.91 & 0.90 & 0.95 & 0.76 & 0.73 & 0.95 & 0.84 & 0.93 & 0.97 & \textbf{0.86 (0.10)}\\
	XDAWN & 0.88 & 0.90 & 0.93 & 0.91 & 0.96 & 0.77 & 0.74 & 0.96 & 0.84 & 0.92 & 0.96 & \textbf{0.86 (0.10)}\\
	\hline
\end{tabular}
\end{table*}

The MDM method offers the best performances among the three methods. Even if the mean improvement is small, the difference is significant (t-test for paired sample; MDM vs. XDAWN : $t(22) = 4.001 $, $p=0.0006$ ; MDM vs. SWLDA : $t(22) = 4.482$, $p=0.0002$). On the contrary, there is no significant difference between the two reference methods (t-test for paired sample; XDAWN vs. SWLDA : $t(22) = 1.172 $, $p=0.26$). 
We can also note that the improvement is higher for subjects with low performances ($0.75 < AUC < 0.9$, e.g., subjects 4, 9, 18, 19 and 21). Overall, the proposed method exhibits an excellent classification accuracy, despite its simplicity.
\newline
\subsubsection{Effect of the size of the training set on performances}
\label{sec:training_size}
\begin{figure}[h]
  \centering\includegraphics[width=1\columnwidth]{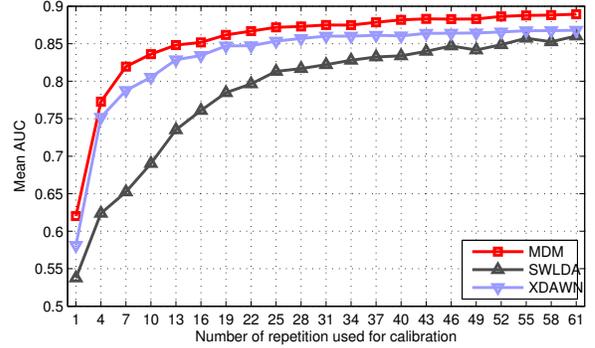}
  \caption{
  Performances of the three methods as a function of the number of trials used for the calibration. 1 repetition = 12 trials, 2 target and 10 non-target.\label{fig:training_size}}
\end{figure}

The size of the calibration data-set is usually a critical parameter influencing dramatically the results.
To evaluate this effect on the MDM algorithm, we calibrate the algorithm using an increasing number of trials from the training set. Figure~\ref{fig:training_size} shows the results of this test.

As expected, the three methods converge to their optimal value. The SWLDA is the method with the slowest convergence; due to the high dimension of the feature space, the model selection needs a lot of data to reach efficiency. On the contrary, the XDAWN method reduces the feature space and therefore needs less data for calibration. Finally, the MDM method shows the best results and the fastest convergence. Thus, the MDM method could be used either to achieve higher performance or to reduce the time spent in the calibration phases, while keeping the same accuracy. As example, in order to obtain a mean AUC of 0.85, only 13 repetitions are needed for the MDM, 22 for XDAWN and 52 for the SWLDA.
\newline
\subsubsection{Effect of the latency and jitter of the evoked potential on performances}
The correct detection of the evoked potentials relies on the fact that the evoked response is perfectly synchronized with the stimulus, i.e., it occurs always at the same time after each stimulus. In practice, a variable delay (jitter) and a fixed delay (latency) is observed, due to technical factors ( Hardware / Software Trigger, screen rendering) as well as physiological factors (fatigue, stress, mental workload).

\begin{figure}[h!]
  \centering\includegraphics[width=1\columnwidth]{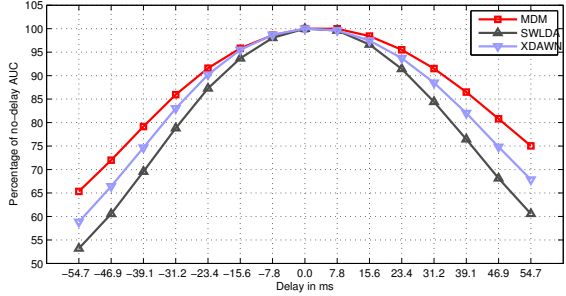}
  \caption{ Effect of the latency on the performances.
  \label{fig:delay}}
\end{figure}

The different classification methods are more or less robust to theses nuisances. These factors are simulated by adding artificial delays on the test data.
To test the effect of latency, the triggers of the test data are shifted with a fixed delay from -55ms to 55 ms. Results are reported in Figure~\ref{fig:delay}.
Performances are normalized with the AUC obtained with a zero delay and subjects 7 and 8 where removed; because of their low performances, close to the chance level, delays have almost no effects on theses two subjects.

\begin{figure}[h!]
  \centering\includegraphics[width=0.9\columnwidth]{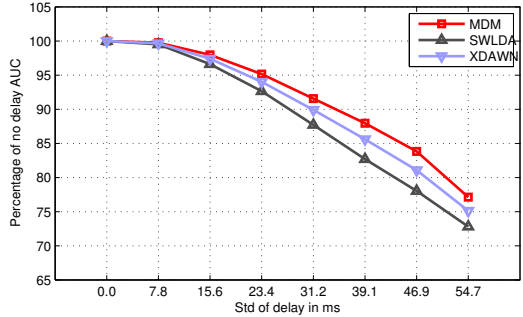}
  \caption{Effect of the jitter on the performances.
  \label{fig:jitter}}
\end{figure}

In Figure~\ref{fig:jitter}, the effect of the jitter is studied by adding a random delay for each triggers of the test data. This delay is drawn from a normal distribution with a zero mean and standard deviation from 0ms to 55ms.

In both cases, the loss of performance induced by the delay is less important for the MDM than for the other methods. Because of the model selection, the SWLDA method is the most sensitive method to the delay. 
\newline
\subsubsection{Effect of $\mathbf{P}_1$ on performances}

The estimation of the super-covariance matrices explained in section~\ref{sec:method} is supervised and requires the average P300 matrix $\mathbf{P1}$. Therefore, it is interesting to see how sensitive the method is regarding to the parameter $\mathbf{P1}$.
To do so, we apply the same test described in section~\ref{sec:canon}, but replacing the subject specific estimation of $\mathbf{P1}$ by the grand average P300 evoked response estimated on other users or using data from another experiment (data from Dataset II).

Results show that there is no statistical difference when a cross-subject $\mathbf{P1}$ (CSP1) or a cross-experiment $\mathbf{P1}$ (CEP1) is used compared to the subject specific $\mathbf{P1}$ ( t-test for paired samples, P1 vs. CSP1 : $t(22) = -0.776$ , $p = 0.45$ ; P1 vs. CEP1 : $t(22) = 0.175$ , $p = 0.86$ ).
These results, particularly meaningful, demonstrate that the critical parameters of the algorithm are the intra-class mean covariance matrices rather than the prototyped evoked responses. Interestingly, it is possible to use an evoked response from another experiment without loss of performances, even if the electrode montage is different. This effect comes from the property of invariance by congruent projection of the Riemannian metric. Thus, the matrix $\mathbf{P1}$ could be estimated in the sources space, with a different electrodes montage or electrodes permutation, and any kind of scaling factor could be applied. We will come back to this interesting property in the discussion section.
\newline
\subsubsection{Cross-subjects performances}
\label{sec:CS}
Results from the previous section show that the supervised estimation of the covariance matrices can be done in an unsupervised way by using the P300 average response from other subjects without loss of performances.
However, calibration data are still required for the estimation of the intra-classe mean covariance matrices. Nevertheless, it is possible to use data from other subjects to estimate the mean covariance matrices as well. Table~\ref{tab:cross_subject} report performances of the different methods for each subject when calibration is done using data from other subjects, according to a leave-one-out procedure.

\begin{table*}[ht!]
\centering
\caption{AUC on test data for the 23 subjects of the dataset I in a subject specific and cross subject (CS) manner.}
\label{tab:cross_subject}
\begin{tabular}{|l|cccccccccccc|}
	\hline
	 & \textbf{1} & \textbf{2} & \textbf{3} & \textbf{4} & \textbf{5} & \textbf{6} & \textbf{7} & \textbf{8} & \textbf{9} & \textbf{10} & \textbf{11} & \textbf{12}\\
	\hline
	MDM & 0.89 & 0.93 & 0.96 & 0.88 & 0.95 & 0.90 & 0.65 & 0.65 & 0.87 & 0.97 & 0.94 & 0.86\\
	CS MDM & 0.71 & 0.92 & 0.82 & 0.80 & 0.90 & 0.72 & 0.73 & 0.64 & 0.87 & 0.80 & 0.88 & 0.86\\
	\hline
	SWLDA & 0.84 & 0.90 & 0.96 & 0.81 & 0.90 & 0.91 & 0.55 & 0.69 & 0.79 & 0.96 & 0.89 & 0.84\\
	CS SWLDA & 0.71 & 0.84 & 0.93 & 0.72 & 0.82 & 0.77 & 0.76 & 0.67 & 0.74 & 0.86 & 0.79 & 0.75\\
	\hline
	XDAWN & 0.89 & 0.90 & 0.96 & 0.85 & 0.91 & 0.90 & 0.56 & 0.66 & 0.79 & 0.95 & 0.83 & 0.86\\
	CS XDAWN & 0.60 & 0.86 & 0.86 & 0.77 & 0.73 & 0.79 & 0.75 & 0.64 & 0.67 & 0.77 & 0.76 & 0.70\\
	\hline
	 & \textbf{13} & \textbf{14} & \textbf{15} & \textbf{16} & \textbf{17} & \textbf{18} & \textbf{19} & \textbf{20} & \textbf{21} & \textbf{22} & \textbf{23} & \textbf{mean (std)}\\
	\hline
	MDM & 0.88 & 0.92 & 0.93 & 0.91 & 0.98 & 0.84 & 0.78 & 0.97 & 0.90 & 0.95 & 0.96 & \textbf{0.89 (0.09)}\\
	CS MDM & 0.89 & 0.77 & 0.91 & 0.89 & 0.95 & 0.72 & 0.77 & 0.84 & 0.76 & 0.85 & 0.84 & \textbf{0.82 (0.08)}\\
	\hline
	SWLDA & 0.87 & 0.91 & 0.91 & 0.90 & 0.95 & 0.76 & 0.73 & 0.95 & 0.84 & 0.93 & 0.97 & \textbf{0.86 (0.10)}\\
	CS SWLDA & 0.86 & 0.72 & 0.86 & 0.86 & 0.94 & 0.67 & 0.79 & 0.87 & 0.74 & 0.84 & 0.83 &\textbf{ 0.80 (0.08)}\\
	\hline
	XDAWN & 0.88 & 0.90 & 0.93 & 0.91 & 0.96 & 0.77 & 0.74 & 0.96 & 0.84 & 0.92 & 0.96 & \textbf{0.86 (0.10)}\\
	CS XDAWN & 0.88 & 0.64 & 0.83 & 0.85 & 0.89 & 0.64 & 0.77 & 0.86 & 0.73 & 0.72 & 0.74 & \textbf{0.76 (0.09)}\\
	\hline
\end{tabular}
\end{table*}

In these circumstances, the MDM method offers the best performances with a mean AUC of 0.82. On the other hand, the XDAWN method displays a really poor performance. This is not surprising since the estimation of spatial filters are known to be very subject specific. Even if the performances of SWLDA are significantly lower than those obtained with the MDM ($t(22) = 1.76,p=0.04$), regarding to its subject specific performances, the SWLDA offers a good generalization across subjects.

We can also notice that for subject 7 the cross subject calibration offers better results as compared to the subject specific calibration. This is generally the case when the calibration data are of low quality, pointing to a general merit of generic initialization.

\subsection{Dataset II}

\begin{figure}[ht!]
  \centering
  \includegraphics[width=1\columnwidth]{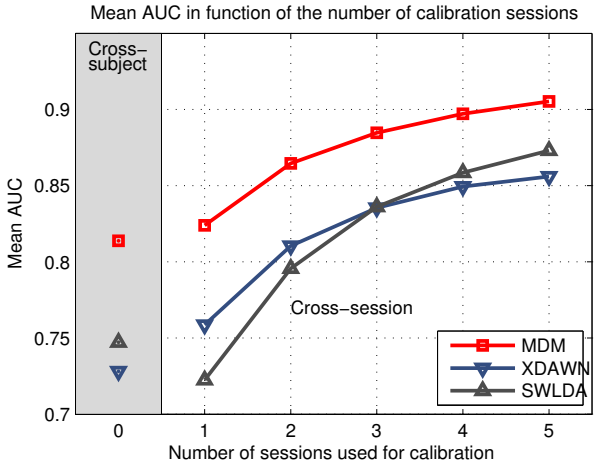}
  \caption{Individual performances accross session for the 4 subjects when an increasing number of sessions are used for calibration.\label{fig:cross-session}}
\end{figure}

For a BCI game such \emph{brain invaders} where the session are generally short (less than 10 minutes) it is not acceptable for the user to do a calibration phase before each game. The second dataset has been recorded to study the generalization of parameters across different sessions. 
Dataset II was recorded in realistic conditions using 16 golden-plated dry electrodes (Fp1-Fp2-F5-AFz-F6-T7-Cz-T8-P7-P3-Pz-P4-P8-O1-Oz-O2) and a USBamp amplifier (Guger Technologies, Graz, Austria). The signal was sampled at 512 Hz. 4 subjects participated to this experiment and made 6 sessions recorded on different days. Each session was composed of a single test phase, without any calibration phase, using cross subjects parameters for the first session and cross sessions parameters for the next sessions.
Offline results are reported in Fig.~\ref{fig:cross-session}. We study first the \emph{cross-subject} performance by initializing the classifier with all the data of the other subjects. Results of number of sessions "0" represent the grand average AUC across all subjects and all sessions. Then we study the \emph{cross-session} performance by initializing the classifier of each subject with an increasing number of his/her own sessions and testing on the remaining sessions. For any $n$ number of sessions used for calibration (x-axis), results represent the grand average AUC obtained on all combinations of the remaining $6-n$ sessions for all subjects.

First, the cross subject classification gives a similar trend as the results obtained on the first dataset. Here the MDM clearly outperform the reference methods. We could also notice that despite we have only four subjects in the database, the cross subject performances are surprisingly good.
Second, the MDM methods shows better cross sessions performances as compared to SWLDA and XDAWN. In this experiment, the sessions are short and contain a small number of trial. For this reason, the SWLDA has difficulty to achieve a high AUC when the number of training sessions is small.
On the average, the same performance could be obtained with only two sessions for the MDM and five sessions for the SWLDA.

\subsection{Dataset III}
Results on dataset I and II demonstrate the good properties of adaptation and generalization of the proposed method. These properties are necessary to obtain an efficient adaptive use of the algorithm.

\begin{figure}[h]
  \centering\includegraphics[width=1\columnwidth]{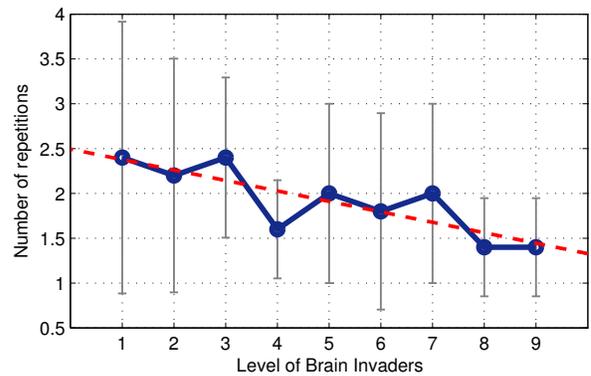}
  \caption{Mean and sd of the number of repetitions needed to destroy the target.
\label{fig:online}}
\end{figure}

Dataset III was recorded in real-life conditions, using the same electrode montage and amplifier as the Dataset II. Sessions were recorded during a live demonstration at the g.tec Workshop, held in our laboratory in Grenoble on March 5th 2013. Five subjects external to our laboratory played the Brain Invaders game for the first time in their life in a noisy environment. They made a single session without calibration using the online adaptive implementation of the MDM described in section~\ref{sec:adaptive}.
Figure~\ref{fig:online} reports online results. The performance is reported in terms of number of repetitions needed to destroy the target alien for each of nine levels of the game.
The adaptive MDM, implemented in python within the open source software OpenVibe~\cite{renard2010openvibe}, was initialized with parameters extracted using the data from the dataset II.
As expected, the adaptation improves the performance (t-test for slope$<0$: $t(7)=-3.82, p=0.007$). On average players needed 2.5 repetitions at the beginning of the game session and only 1.5 for the last level. In addition, the variability decreases, and  typically for the last two levels less than two repetitions only suffice to destroy the target. This is a very good result as compared to the state-of-the-art.

\section{Conclusion}
\label{sec:conclusion}

We have proposed a new classification method for ERP-based BCI that increases performance, reduces the number of data needed for the calibration and features good generalisation across sessions and subjects. In addition, this new method proves more robust against fixed and variable delays on the trigger signals as compared to state-of-the-art methods. An adaptive implementation has been deployed on our real-life experiments allowing the user to play without any calibration (plug\&play) while keeping a high accuracy. The Brain Invaders and the OpenViBE scenarios we use are open-source and available for download\footnote{http://code.google.com/p/openvibe-gipsa-extensions/}.
\section{Discussion}
In this work a very simple classifier (MDM) has been used in order to facilitate the adaptive implementation. In our previous works on motor imagery more sophisticated classification algorithms has been developed, leading to higher performances as compared to the MDM. In~\cite{barachant2010riemannian}, a geodesic filtering was applied before the MDM in order to improve class separability. In~\cite{barachant_neurocomp_2013}, a SVM-kernel dedicated to covariance matrices was defined and another way of dealing with inter-session variability was proposed. These methods give promising results but show less stability in a cross-subject context. In future works we will investigate whether more complex classification methods based on information geometry can be used in an adaptive context.

With the new definition of covariances matrices proposed in this article the Riemannian geometry framework can now be applied on all the three main BCI paradigms (SMR, ERP, SSEP) with only minor modifications. A single implementation of the classification method can be used for the different paradigms, which is very convenient in terms of software development and maintenance.
Besides the Riemannian frameworks, the new proposed estimation of covariance matrix for ERP brings permeability between states-of-the-art methods since the wide majority of signal processing alorithms for SMR-based BCIs rely on the manipulation of covariance matrices. For example, the well known CSP spatial filtering and all its variations can now be applied for ERPs.

Finally, the property of invariance of the metric gives great freedom in the choice of the prototyped ERP matrix $\mathbf{P1}$. This matrix could be drawn by hand, or generated with a wavelets basis, allowing, for example, to adapt the system to different sampling frequency. 

\change{
However, this approach shows some limitations when the number of channels increase. If the number of channel is greater than the number of time sample in the trial, the covariance matrix is ill-conditioned and the computation of the Riemannian distance becomes impossible. In this context, a regularized estimate of the covariance matrix must be used, or if it is possible, a higher sampling frequency. In addition, if the number of channel is too big ($>64$), the estimation of the mean covariance matrices might not converge. In this case, consider to use a smaller number of channel or apply a subspace reduction (PCA) to decrease the dimensionality of the problem.
}
\bibliographystyle{IEEEtran}
\bibliography{RiemannP300.bbl}
\begin{IEEEbiography}[{\includegraphics[width=1in,height=1.25in,clip,keepaspectratio]{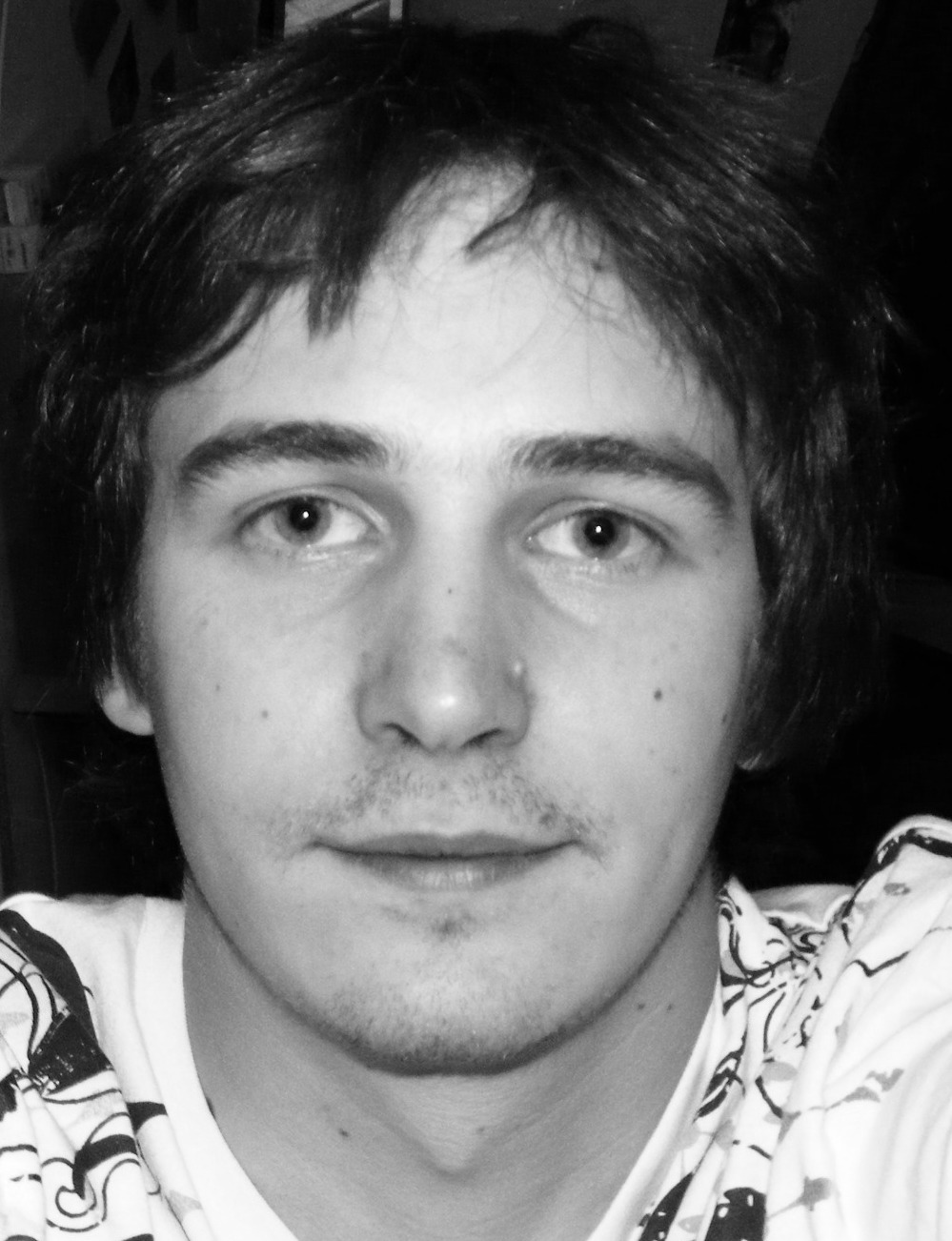}}]{Alexandre Barachant}
 was born in Chateauroux, France, in 1985. 
In 2012, he recieved the Ph.D. degree in signal processing from the Grenoble University, Grenoble, France.
He has worked at CEA-LETI during his Ph.D. on the topic of brain computer interfaces. He his actually a post-doc fellow at the Centre National de la Recherche Scientiﬁque (CNRS) in the GIPSA Laboratory, Grenoble, France. His current research interest include statistical signal processing, Riemannian geometry and classification of neurophysiological recordings with applications to brain computer interfaces.
\end{IEEEbiography}
\begin{IEEEbiography}[{\includegraphics[width=1in,height=1.25in,clip,keepaspectratio]{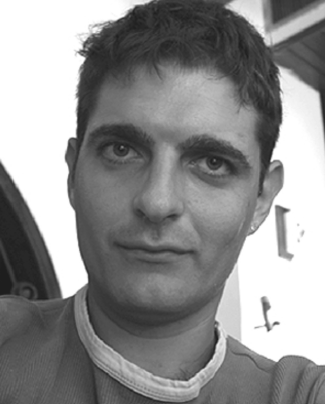}}]{Marco Congedo}
 obtained the Ph.D. degree in Biological Psychology with a minor in Statistics from the University of Tennessee, Knoxville, in 2003. From 2003 to 2006 he has been a post-doc fellow at the French National Institute for Research in Informatics and Control (INRIA) and at France Telecom R\&D, in France. Since 2007 Dr. Congedo is a Research Scientist at the "Centre National de la Recherche Scientifique" (CNRS) in the GIPSA Laboratory, Grenoble, France.
 Dr. Congedo has been the recipient of several awards, scholarships and research grants. He is interested in basic human electroencephalography (EEG) and magnetoencephalography (MEG), real-time neuroimaging (neurofeedback and brain-computer interface) and multivariate statistical tools useful for EEG and MEG such as inverse solutions, blind source separation and Riemannian geometry. Currently he is Academic Editor of journal PLoS ONE.
\end{IEEEbiography}
\vfill
\end{document}